# Gaussian Three-Dimensional kernel SVM for Edge Detection Applications

Safar Irandoust-Pakchin[1*] - Aydin Ayanzadeh[2]
Siamak Beikzadeh[3]

[1*] Department of Applied Mathematics Faculty of Mathematical Sciences, University of Tabriz, Iran
[2] Computer Science Department, Faculty of mathematical sciences, university of Tabriz, Tabriz, Iran
[3] Computer Science Department, Faculty of mathematical sciences, university of Tabriz, Tabriz, Iran

ABSTRACT

*This paper presents a novel and uniform algorithm for edge detection based on SVM (support vector machine) with Three-dimensional Gaussian radial basis function with kernel. Because of disadvantages in traditional edge detection such as inaccurate edge location, rough edge and careless on detect soft edge. The experimental results indicate how the SVM can detect edge in efficient way. The performance of the proposed algorithm is compared with existing methods, including Sobel and canny detectors. The results shows that this method is better than classical algorithm such as canny and Sobel detector.*

## 1.  INTRODUCTION

Edges in a digital image provide important information about the objects contained within the image since they constitute the boundaries between the objects in the image [1]. Edge detection has the lot of application in the image processing and one of the fundamental and key issue in the image analysis, pattern recognition, computer vision, medical imaging[2]. In order to extract the contour of an object in an image. We must detect the complete information of the edges, so edge detection is an indispensable part of image processing. These can be used as final result or as intermediate information for further interpretations, such as segmentation, face recognition, object recognition, 3D reconstruction, defect detection on mechanical parts, tracking, image retrieval or stereo vision, etc. Detecting edges is very useful in a no of contexts. For example in a typical image understanding task such as object identification, an essential step is to segment an image into different regions corresponded to different objects in the scene. Edges are sets of pixels in the image regions with sharp intensity changes and correspond to visible contour features of objects in an image. Normally, edge detection is a process that inputs a grayscale image and then outputs a binary image to indicate the edges of objects. [3-6].

Many edge detection methods have been proposed in the last decade. Most of them are based on digital differential methods such as Sobel, Canny [7], Prewitt, Roberts, Laplacian operators and so on use different discrete approximations of the derivative function.They are applied medicine and are used for protection of important structural properties, images and etc. traditional methods include  high sensitivity to noise and inability to discriminate edges versus textures. The most usual classical methods search for several way to perform an approximation to the local derivation and they mark the edge by searching the maximum of these derivation. The main idea in this work is to train the computer to recognize the presence of edges into an image. In order to perform this idea we use the Support Vector Machines with Three-dimensional Gaussian kernel for our classification problems. These algorithms can detect the local features of the signal at dissimilar frequency.

In the last decade, SVM has proven to be an effective method in the field of machine learning. SVM is a popular algorithm for classification learning. SVM has been applied to solve a variety of practical problems [8]. It has been widely used in various applications, including handwriting [9], digits recognition, natural language understanding, time series classification gene selection, image retrieval and etc.

2. **SVM**

SVMs have been developed in the reverse order to the development of neural networks (NNs). SVMs evolved from the sound theory to the implementation and experiments, while the NNs followed more heuristic path, from applications and extensive experimentation to the theory. Support vector machines (SVM) is a learning method.it mostly use for classification. In the elementary design of an SVM classifier, the bounding planes of each class are considered and the distance between the bounding planes is defined as the margin. The aim of SVM is to find the optimal separation boundary via maximizing the margin between classes. In real problems that we are involve with them the separator is more complicated and it is not always linearly and it must our input vectors go to the higher dimension space to find a to separating classes[10]. Through the process for having an optimal separator we faced a primal problem.

$$\min \frac{1}{2} w^T w \qquad (1)$$
$$\text{St y: } (w^T x_i + b) \geq 1$$

To solve that we used an alpha ($\alpha \geq 0$) after all this we faced a QP problem and there are many algorithms to solve that in this case we used interior-point-convex algorithm. According to last researches we can use smaller QP [11]. These methods only suggest memory requirements linear in the number of training examples, they may still need a long training time because the time complexity is closely related to the number of iterations. To solve that in this case we used interior-point-convex algorithm.

A variety of edge detection have been presented in image processing association although seemingly very various, they all share the same goal to keep meaningful edge and remove less-meaningful ones. Give a training dataset $\{(x_i, y_i)\}_{i=1}^{N}, x_i \in R^n, y_i \in \{-1, 1\}$ where $x_i$ is the input vector with known binary target $y_i$, the original SVM classifier (vapnik,1998), satisfies the following condition or equivalently:

$$y_i[W^T \emptyset(x_i) + b] - 1 \geq 0, i = 1, \ldots, N \qquad (2)$$

Where $\emptyset: R^n \to R^m$ is the feature map mapping the input space to usually high dimensional feature space where the data points become linearly separable by a separating hyperplane defined by the pair of classification function is then given by:

$$f(x) = sign\{W^T \emptyset(x) + b\} \qquad (3)$$

It is usually unnecessary to calculate with the feature map and one only needs to work instead with a kernel function in the origin AI(Artificial Intelligence)space given by $k(x_i, y_i) = \emptyset(x_i)^T \emptyset(x_j)$ in order to allow for the violation of EQ, we introduce slack variables $\varepsilon_i$ such that :

$$y_i[w^T \phi(x_i+b)] \geq 1-\epsilon_i, \quad \epsilon_i > 0, \quad i=1,2,...,N \tag{4}$$

and consider the following minimization problem:

$$\min_{w,b,\xi} \quad J(w,b,\xi) = \frac{1}{2}\|w\|^2 + C\sum_{i=1}^{N} \xi_i \tag{5}$$

$$\text{s.t.} \quad y_i[w^T \phi(x_i) + b] \geq 1 - \xi_i \tag{6}$$

$$\xi_i > 0, \quad i = 1,\ldots,N, \quad C > 0 \tag{7}$$

### 3. EDGE DETECTION TRAINING WITH THREE-DIMENSIONAL SVM

In this section we present a way to detect edges by using SVM classification. In this case, the decision needed is between "the pixel is part of an edge" or "the pixel is not part of an edge" for this case we need some information of our image to extract metric or property for our process. We consider the dark zone and bright zone using difference of luminous intensity and position of each pixel. Thus we create our tridimensional vectors for our SVM inputs. Our SVM details got a little change because the SVM that is traditional has maximum two-dimensional input but we are trying to improve it to support tridimensional inputs and the kernel that we used is Gaussian kernel with improved to tridimensional inputs. The Gaussian kernel is a popular kernel function used in various kernelized learning algorithms. Quadratic programming is the problem of finding a vector α minimizes a quadratic function, possibly subject to linear constrain:

$$\begin{cases} \min \quad \frac{1}{2}\alpha^T H \alpha + f^T \alpha \\ S.T \quad \sum_i \alpha_i y_i = 0 \\ \quad \alpha_i \geq 0 \end{cases} \tag{8}$$

$$Gaussian\ Kernel = \exp\left(\frac{1}{-2\sigma^2}\|x_i - x_j\|^2\right) \tag{9}$$

H is related to our quadratic programming problem in Two-Dimensional mode

$$H_{ij} = y_i \cdot y_j \cdot x_i^T x_j \rightarrow H = y_i \cdot y_j \ker(x_i, x_j) \tag{10}$$

We extend it to tridimensional mode with this method:

$$H = y_i\, y_j\, y_z\, ker\,(x_i, x_j, x_z) \tag{11}$$

$$k(x_1, x_2, x_3, \ldots, x_n) \tag{12}$$

We define x, y and z as center of gravity with this formulae:

$$x = \frac{\sum_{i=1}^{n} x_i}{n} \tag{13}$$

$$y = \frac{\sum_{i=1}^{n} y_i}{n} \tag{14}$$

$$z = \frac{\sum_{i=1}^{n} z_i}{n} \tag{15}$$

And calculate distance from vector to center of gravity as radius square:

$$\text{Radius square} = (X - x)^2 + (Y - y)^2 + (Z - z)^2 \tag{16}$$

Finally the proposed Tridimensional Gaussian kernel for SVM is :

$$Ker = \exp(\frac{1}{2\sigma^2}\, Radius\ Square) \tag{17}$$

The vector formed are used as inputs to the SVM training and return support vector machine. This SV (support vector) space gives critical zones that called edge. The pixel considered as edge are those into each image that are in the border between bright and dark zones the image used to train the SVM are shown in Fig1.they are images created by trying to obtain a good model for the detection. The only edges used in the training are vertical, horizontal and diagonal ones as show in Fig1 and we expect that the other edges will be generalized by the SVM.

The dark and bright zones are heterogeneous but the intensity at each pixel is random value. By using these random value we try to simulate the heterogeneous surface into real image. A value that must be set in the training process is the mean difference between dark and bright zones, since this parameter control the detector more sensible and a greater one reduce this sensibility.

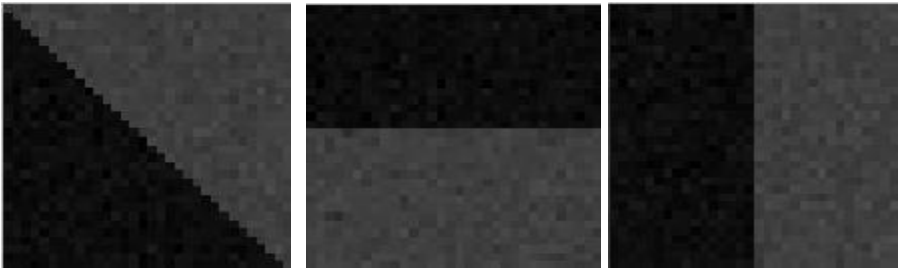

Fig. 1. image training for edge detection first column is a diagnal edge and column 2 and 3 is horizontal and vertical edge

When we apply the trained SVM to an image, a value for each pixel into the image is obtained. In ideal condition this value must be a positive or negative value near 1. Use the value obtained and say that there is a quantized change between "no edge" (an area that does not have a significant change occurs in some physical aspect of an image) and edge. Then the value obtain indicates a probability of being an edge or not. After the process above we must decide the pixels that are considered as edges. And we can print color in SV space and SV candidate the same color that the support vector is near the class that with little high probability is belong.

## 4. RESULTS

To explore the utility and demonstrate the efficiency of the proposed edge detection approach, we carry out computer experiments on grayscale images. A fixed threshold is used in the experiments for simplicity, although there are adaptive thresholding techniques that could be implemented. We set the optimize value in our experiment and obtain an efficient results in simulation according to Fig 2.

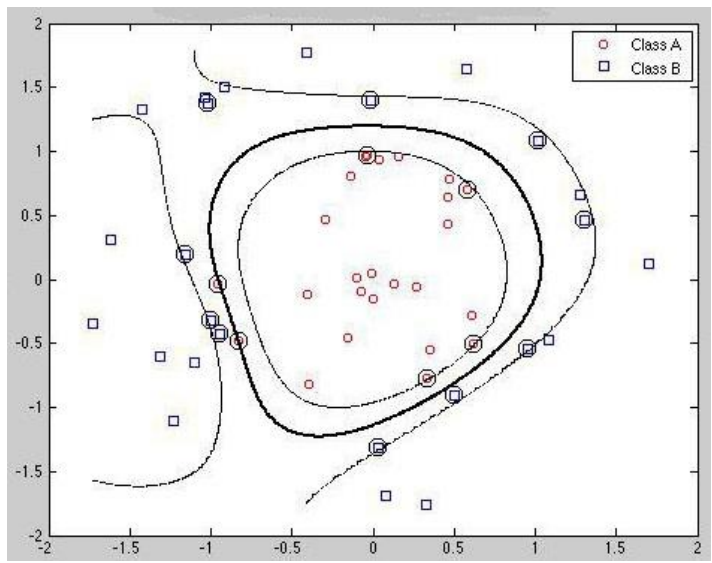

Fig.2. svm classification with propose method in optimization mode with c=10 and σ =0.6

A list of the images used in the present study is plotted in Fig.3. The cameraman, house and tire. These are our tested image that we implemented our new method on it. The resolution of all images is 8-bit per pixel and the size of them are 256*256 . The computer experiments are conducted to test the proposed approach. The experiments are designed to compare the ability of the proposed method on extracting edges from clean image with the standard Canny and Sobel  detector. From the result of this simulation according Fig4  we find out that SVM has higher classification accuracy in the edge detecting and this method is more sensitive than other existing edge detectors methods. We can see the impact of the performance of proposed SVM detectors . The proposed algorithm effectively detects more fine and fewer spurious structures than the Canny algorithm. In first row of Fig4, canny algorithm is more inaccurate on edge detection and create excessive edge in some location of image.

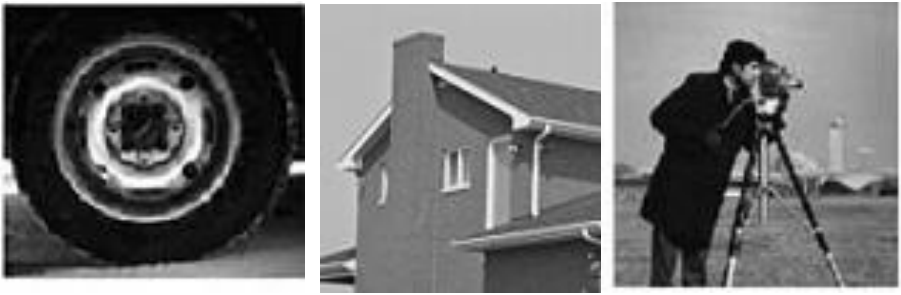

Fig.3. A collection of sample images where Tire , CameraMan, House is of size is 128*128

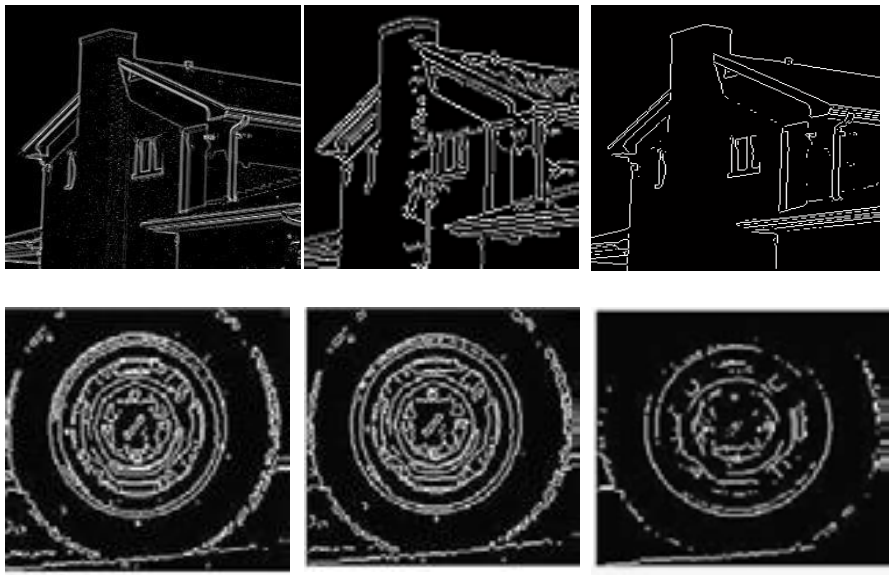

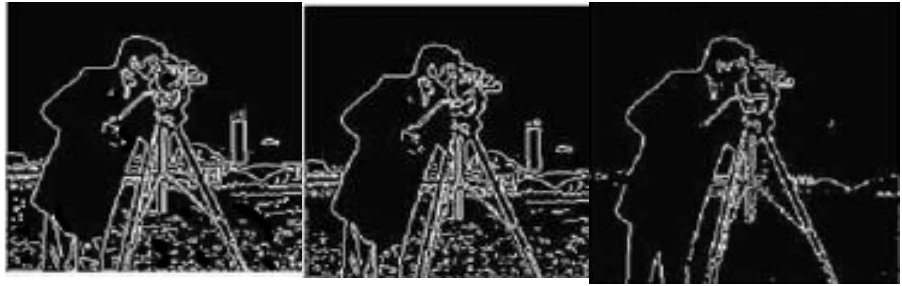

Fig.4. Edge images obtained by different detectors.column 1 is obtain by the proposed approach. Column 2 and 3 is respectively tested by standard Canny and Sobel detector

To demonstrate the efficiency of the proposed edge detector the elapsed times are also counted, the statistics of the elapsed time for different methods are shown in Table 1. From Table 1, the proposed edge detectors are shorter than the Canny in process time, which means it is much more efficient. All the experiments were performed on the Intel Core i3 4030U 1.9 GHZ laptop. For image Cameraman of size 256*256, the proposed algorithm needs only about 1.02 while the Canny needs about 1.22s. The proposed method in Cameraman approximately 0.2s faster than canny detector in process time. The same results are demonstrated by the other image edge detection experiments.

Tabel1. The statistics of the process time for different edge detectors

| Tested image | Proposed Method(s) | Canny(s) | Sobel(s) |
|---|---|---|---|
| House | 0.83 | 0.94 | 0.19 |
| Tire | 0.71 | 0.87 | 0.22 |
| Cameraman | 1.02 | 1.22 | 0.27 |

## 5． CONCLUSION

This work shows how the SVM performs the edge detection in presence in an efficient way and given good results. A number of gradient operator are obtained from the SVM with Gaussian kernel, and the new edge detector, based on tridimensional input for Gaussian kernel. The performance of the proposed algorithm is compared with Canny and Sobel detectors. Experiments on images have been tested in MATLAB and was implemented with Gaussian kernel without any library such as Lib-SVM and it shows that the proposed edge detector more fine and fewer spurious structures than the Sobel and Canny algorithm.

## REFERENCES

[1] Yuksel ME. (2007). Edge detection in noisy images by neuro-fuzzy processing. AEU – Int J Electron Commun
[2] Duan S, Hu X, Wang L, Gao S, Li C. (2013). Hybrid memristor/RTD structure-based cellular neural networks with applications in image processing. Neural Comput Appl 25(2). p 162-164
[3] Q. Sun,Y.Hou,Q.Tan,C.Li, (2014). Shaft diameter measurement using a digital image, Opt. Lasers Eng.55. p183–188.

[4] Q. Sun,Y.Hou,Q. Tan,C.Li,M.Liu. (2014), A robust edge detection method with sub-pixelaccuracy, Optik125. p 3449–3453.

[5] Lopez-Molina C, De Baets B, Bustince H. (2011). Generating edge images from gradient magnitudes. Comput Vision Image Understand.

[6] L. Fan, F. Song, S. Jutamulia, (2007) . Edge detection with large depth of focus using transform,Opt.Commun.270.

[7] J.F. Canny.(1986). A computational Approach to Edge Detection. IEEE Trans.Pattern Anal. March. Intell. P 679-698

[8] Yunqian Ma, Guodong Guo, (2014). Support Vector Machines Applications , Springer, preface section p 2-10

[9] X.-X. Niu, C. Y. Suen, (2012). A novel hybrid cnn–svm classifier for recognizing handwritten digits, Pattern Recognition 45 (4) p 1318–1325.

[10] Lipo Wang (2005). Support Vector Machine Theory and Applications, Springer, p 2-35

[11] Boyang Li, Qiangwei Wang , Jinglu Hu, (2011). Fast Svm Training using Edge Detection on Very Large Dataset. Published by John Wiley & Sons, Inc.